\documentclass[preprint,12pt]{elsarticle}
\usepackage{lineno}
\modulolinenumbers[1]
\usepackage{geometry}
\usepackage{graphicx}
\usepackage[usenames,dvipsnames]{color}
\usepackage[english]{babel}
\usepackage{epstopdf}
\usepackage{latexsym}
\usepackage{url}
\usepackage{amstext}
\usepackage{amssymb}
\usepackage{amsmath,amsthm}
\usepackage{pifont}
\usepackage{ulem}
\usepackage[pagebackref=false]{hyperref}
\hypersetup{
    colorlinks=true,
    linkcolor=blue,
    filecolor=magenta,
    urlcolor=blue,
}

\biboptions{sort&compress}
\begin{document}

\begin{frontmatter}

\title{Machine Learning for Complex Systems Dynamics: Detecting Bifurcations in Dynamical Systems with Deep Neural Networks}

\author[inst1,inst2]{Swadesh Pal}
\ead{spal@wlu.ca}
\author[inst1,inst3]{Roderick Melnik}
\ead{rmelnik@wlu.ca}

% \cortext[cor1]{Corresponding author}
\address[inst1]{MS2Discovery Interdisciplinary Research Institute, Wilfrid Laurier University, Waterloo, Canada}
\address[inst2]{Department of Mathematics, SRM Institute of Science and Technology, Kattankulathur, India}
\address[inst3]{BCAM - Basque Center for Applied Mathematics, Bilbao, Spain}

\begin{abstract}
Critical transitions are the abrupt shifts between qualitatively different states of a system, and they are crucial to understanding tipping points in complex dynamical systems across ecology, climate science, and biology. Detecting these shifts typically involves extensive forward simulations or bifurcation analyses, which are often computationally intensive and limited by parameter sampling. In this study, we propose a novel machine learning approach based on deep neural networks (DNNs) called equilibrium-informed neural networks (EINNs) to identify critical thresholds associated with catastrophic regime shifts. Rather than fixing parameters and searching for solutions, the EINN method reverses this process by using candidate equilibrium states as inputs and training a DNN to infer the corresponding system parameters that satisfy the equilibrium condition. By analyzing the learned parameter landscape and observing abrupt changes in the feasibility or continuity of equilibrium mappings, critical thresholds can be effectively detected. We demonstrate this capability on nonlinear systems exhibiting saddle-node bifurcations and multi-stability, showing that EINNs can recover the parameter regions associated with impending transitions. This method provides a flexible alternative to traditional techniques, offering new insights into the early detection and structure of critical shifts in high-dimensional and nonlinear systems.
\end{abstract}


\end{frontmatter}

\section{Introduction}

Complex systems in nature and society often exhibit sudden, dramatic changes in their behaviour, referred to as critical transitions or tipping points. These phenomena, ranging from ecosystem collapses and abrupt climate changes to financial crashes and the onset of neurological diseases, are typically driven by gradual changes in underlying parameters, yet result in nonlinear, discontinuous shifts in system dynamics. In addition, these abrupt changes are characterized by the appearance, disappearance, or qualitative shift in equilibrium points or attractors \cite{scheffer2001}. Identifying such transitions is crucial for early warning and prediction across various scientific and engineering disciplines \cite{rietkerk2004, scheffer2003, donangelo2010, wu2024}. Detecting early warning signals of such transitions, such as critical slowing down or rising variance, has become a central research theme. The integration of ordinary differential equation (ODE) models with data-driven techniques, including machine learning and real-time sensing, has opened new avenues for monitoring and predicting catastrophic events in diverse systems. These models not only deepen our theoretical understanding but also inform practical strategies for resilience and intervention in vulnerable systems.

Traditional methods for analyzing critical transitions rely on linear stability analysis, continuation techniques, and extensive numerical simulations \cite{grziwotz2023}. In addition, the mathematical characterization of these transitions is achieved through bifurcation analysis of nonlinear, parametrized ODEs. These approaches become particularly expensive when applied to high-dimensional problems, especially when the system is nonlinear and has multiple equilibria with respect to a bifurcation parameter. Moreover, identifying bifurcation diagrams in such models requires fine-grained sampling of the parameter space and repeated integrations of the full system, making them impractical for real-time or large-scale applications.

In recent years, machine learning (ML) has emerged as a promising alternative to traditional numerical approaches, offering tools for dimensionality reduction, pattern recognition, and nonlinear function approximation \cite{sarker2021, pichler2023}. For instance, a deep learning classifier provides early warning signals for five local codimension-one discrete-time bifurcations \cite{bury2023}. Additionally, researchers trained deep learning models on time series data from prototypical systems, incorporating both time-varying equilibrium shifts and stochastic noise \cite{huang2024}. In this work, we propose and explore a framework that leverages ML techniques, such as deep neural networks (DNNs), to reduce the computational burden of finding equilibrium points and tracking bifurcations. Traditional numerical methods for computing equilibria often rely on parameter sweeps or root-finding algorithms, where the steady-state equations are solved directly for fixed parameter values. While such methods can be adapted to construct bifurcation diagrams using DNNs, they become challenging when the system exhibits multiple equilibrium points for a single parameter value. To address this, we introduce a reverse methodology: instead of sweeping over parameters, we provide a range of candidate values representing potential equilibrium states and then identify the corresponding parameter values for which these equilibria occur. Since this approach uses the possible equilibrium points as input to the DNN, we refer to it as the equilibrium-informed neural networks (EINNs) method. By systematically exploring a continuous space of equilibrium states, EINNs can generate bifurcation diagrams and detect critical transitions such as saddle-node or Hopf bifurcations without relying on exhaustive simulations. The core idea is to use deep learning paradigms to model the equilibrium-to-parameter mapping, detect early-warning signals of bifurcations, and potentially classify the type and stability of the transitions.

Applications of this methodology are broad. In ecological systems, this can represent a sudden collapse of a population once the environment deteriorates past a tipping point, even if the degradation is gradual \cite{scheffer2003, may1977, dakos2008, carpenter2015, mumby2007, pal2024}. Beyond ecology, ODE models exhibiting catastrophic shifts have been applied in numerous fields. In climate science, they describe abrupt transitions such as the shutdown of ocean circulation systems \cite{rahmstorf2006}. In neuroscience, they help model seizure onset or the collapse of functional connectivity in neurodegenerative diseases \cite{fitzhugh1961}. In economics, ODEs capture market crashes and financial instability \cite{smug2018}, while in engineering, they are used to study failure cascades in complex infrastructures.

The remainder of this article is organized as follows. In Section \ref{SE2}, we introduce the ODE systems where critical transitions occur. Section \ref{SE3} presents the ML frameworks and how the EINNs approach works. In addition, we have applied the EINNs approach across various models, ranging from a single equation to systems of equations. Finally, we conclude in Section \ref{SE4} with remarks on the role of ML in advancing the science of complex systems, current limitations, and directions for future research.

\section{Problem Setup}{\label{SE2}}

ODE models are fundamental tools for describing the dynamics of systems that evolve continuously over time. In ecology, ODEs are widely used to model the population dynamics of species, the interactions between species, and the influence of environmental drivers on ecosystem stability. These models provide a framework to understand not only the gradual changes in state variables, such as population size or nutrient concentration, but also the potential for catastrophic shifts - sudden, often irreversible changes in system behaviour resulting from smooth variations in external conditions. A key mathematical mechanism underlying catastrophic shifts is the saddle-node bifurcation, where two fixed points (a stable equilibrium and an unstable one) collide and annihilate each other as a control parameter passes a critical threshold \cite{scheffer2001}. In this context, we focus on a system of parametrized nonlinear ordinary differential equations of the general form:
\begin{equation}{\label{TM}}
        \frac{d\mathbf{u}}{dt} = \mathbf{F}(\mathbf{u}; \lambda),~~\mathbf{u}\in\mathbb{R}^{n}, \lambda\in\mathbb{R},
\end{equation}
with an initial condition $\mathbf{u}(0) = \mathbf{u}_{0}\in\mathbb{R}^{n} (n\geq 1)$. Here, $\mathbf{F}(\mathbf{u};\lambda)$ is a vector-valued nonlinear function in $\mathbb{R}^{n}$ parametrized by the parameter $\lambda$ representing external environmental or anthropogenic drivers (e.g., temperature, harvesting rate, nutrient input). A critical shift occurs at a parameter value $\lambda = \lambda_{c}$ if the system undergoes a bifurcation at $\lambda_{c}$, resulting in a qualitative change in the set or stability of equilibrium points or attractors. That is, there exists a neighbourhood $\mathbf{U}\subset \mathbb{R}^{n}$ and an interval $I\subset \mathbb{R}$ containing $\lambda_{c}$ such that:
\begin{itemize}
    \item For $\lambda < (>) \lambda_{c}$, the system admits an attractor $\mathbf{A}_{1}\subset \mathbf{U}$ (e.g., a stable equilibrium or limit cycle),
    \item For $\lambda > (<) \lambda_{c}$, either ceases to exist or loses stability, being replaced by a topologically distinct attractor $\mathbf{A}_{2} \subset \mathbb{R}^{n}$ that is different from $\mathbf{A}_{1}$,
    \item The transition from $\mathbf{A}_{1}$ to $\mathbf{A}_{2}$ is a significant change (i.e., discontinuous in the attractor structure), often associated with hysteresis or a loss of resilience.
\end{itemize}
Different types of bifurcations leading to critical shifts; however, we focus only on the saddle-node bifurcations in this work. This setup encapsulates a wide range of problems, such as the kinetic equations in ecology and biology. We first discuss the ML method for a single-equation model, and then we discuss the approach for systems involving two or more equations. 

\section{EINNs Approach and its Applications}{\label{SE3}}

This work aims to lay the foundations for predicting possible critical transitions based on saddle-node bifurcations in mathematical models with the help of deep learning approaches. In this section, we present the EINNs approach in detail and demonstrate its effectiveness across a range of dynamical systems. We begin by outlining the general architecture of the EINNs framework, including network design, training strategies, and loss function construction. We then apply the method to representative models that exhibit complex bifurcation behaviour, showcasing how EINNs can be used to construct bifurcation diagrams, detect critical thresholds, and identify qualitative changes in system dynamics. However, EINN's method does not detect the abrupt shifts in time series or calculate early warning indicators linked to critical slowing down.

\subsection{Single-equation dynamical models}

Single-equation models provide fundamental building blocks for complex models involving complex interactions. Environmental factors influencing ecosystems, such as climate change, nutrient enrichment, pollution, groundwater depletion, habitat fragmentation, or biodiversity reduction, often evolve gradually, sometimes even linearly. In some ecosystems, these slow changes lead to smooth, continuous shifts in ecological states. However, other systems may remain unchanged across various conditions, only exhibiting noticeable shifts as environmental pressures near a critical threshold. In a different scenario, an ecosystem’s response may exhibit a ``folded" behaviour, where the system can exist in one of two distinct stable states under the same external conditions. These states are separated by an unstable equilibrium, which serves as the tipping point between their respective basins of attraction. For a single equation model, we write the system (\ref{TM}) with a parameter $\lambda$ as: 
\begin{equation}{\label{SSM}}
    \frac{du}{dt} = f(u;\lambda),
\end{equation}
with the initial condition $u(0) = u_{0}$. 

Here, we discuss the cases where critical shifts occur through the generation or disappearance of a stable equilibrium point in a system. We need to identify all equilibrium points across a range of $\lambda$ values to do this. The traditional approach tracks the equilibrium points by solving the nonlinear equation $f(u;\lambda)=0$ for each fixed $\lambda$, and then plotting all the solutions of it in a diagram, illustrating how the number of equilibria changes with $\lambda$ \cite{kuznetsov1998,strogatz2024}. In this process, the critical transition thresholds can be detected by finding the values for $\lambda$ for which the appearance or disappearance of the equilibrium points happens in the diagram. Finding the equilibrium points of the equation $f(u;\lambda)=0$ across a range of values for the parameter $\lambda$ is a non-trivial task, particularly in systems where multiple equilibrium points exist. Traditional numerical methods can struggle to capture all possible fixed points, especially in high-dimensional or nonlinear settings. However, the following machine learning approach can be employed to construct such a bifurcation diagram for a range of $\lambda$ values, even if finding the solutions of the nonlinear equation $f(u;\lambda)=0$ is challenging.

DNNs are powerful tools for approximating complex, nonlinear functions under certain regularity conditions. In our work, we aim to leverage DNNs to approximate the solutions of the parameterized equation $f(u;\lambda)=0$ over a range of values for the parameter $\lambda$, and we denote the solutions as $u_{*}$. Specifically, we are interested in identifying equilibrium points, i.e., values $u=u_{*}$ that satisfy this equation for varying $\lambda$, which may represent the governing systems' inputs, environmental conditions, or control parameters, depending on the application. 

Generally, classical approaches using DNNs to find equilibrium points involve fixing the value of $\lambda$ and training the network to produce one or more solutions for $u$ that minimize the residual error $|f(u;\lambda)|^{2}$. In such cases, the architecture of the DNNs (particularly the number of output nodes) plays a crucial role in determining the number of distinct solutions that can be represented, and the network may converge to the same solution at multiple output nodes, resulting in duplicate equilibria. Additionally, it may happen that the number of output nodes is less than the possible number of equilibria of the system. This redundancy often reflects the limitations of the network training. This is especially valuable in systems with multistability or bifurcation behaviour, where multiple equilibria coexist for the same parameter values.

\begin{figure}
\centering
\includegraphics[width=0.45\textwidth]{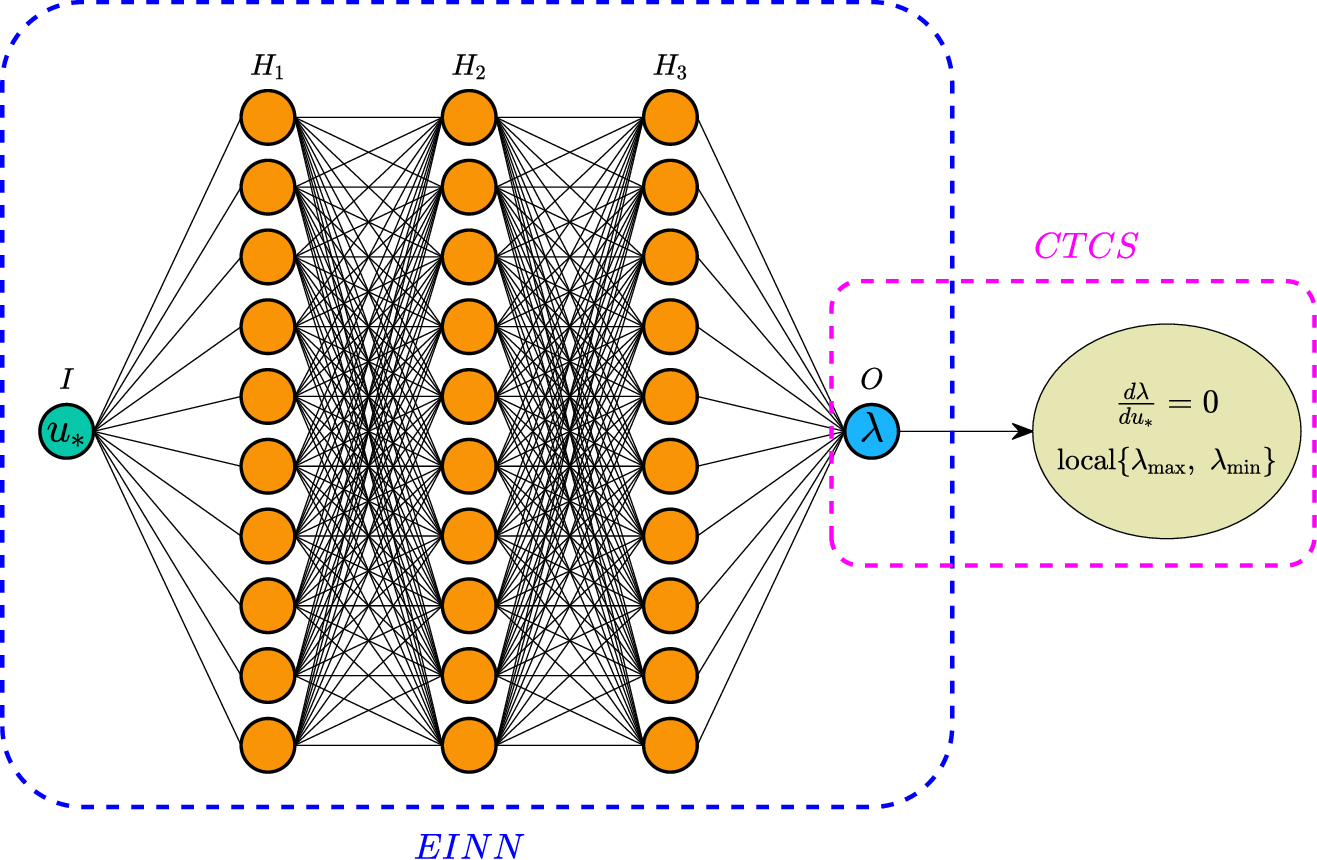}
\caption{ (Color online) Schematic diagram of the DNN-based inversion approach. The input and output layers are denoted by $I$ and $O$, respectively, while $H_{1}$, $H_{2}$, and $H_{3}$ represent the hidden layers of the DNN. CTCS refers to the critical thresholds for catastrophic shifts. } 
\label{FigN1}
\end{figure}

To overcome the limitations of the traditional approach, where fixing the parameter $\lambda$ may only yield a limited or redundant set of solutions, we propose an alternative, reverse methodology. Instead of holding $\lambda$ constant and searching for corresponding solutions $u_{*}$, we begin by selecting a range of candidate values $\{u_{*}^{j}\}_{j=1}^{N}$, which are potential equilibrium states of the system. Here, $N$ denotes the total number of collocation points sampled from the prescribed range of $u_{*}$ that are used during the training process. The objective is then to determine the set of values $\{\lambda^{j}\}_{j=1}^{N}$ that make each chosen $u_{*}^{j}$ a valid solution to the equation $f(u^{j};\lambda^{j})=0$. To achieve this, we formulate an optimization problem in which the DNN is trained to minimize the mean squared residual error of the function, defined as: 
\begin{equation}{\label{SEOPTM}}
    MSE = \frac{1}{N}\sum_{j=1}^{N} |f(u_{*}^{j};\lambda^{j})|^2.
\end{equation}
This framework allows the DNN to learn the mapping from a given candidate solution $\{u_{*}^{j}\}_{j=1}^{N}$ to the set of parameter values $\{\lambda^{j}\}_{j=1}^{N}$ that make each $u_{*}^{j}$ an equilibrium point of the system corresponding to the parameter $\lambda^{j}$. In this approach, we provide the potential equilibrium candidates as inputs and train the neural network accordingly, which we refer to as the EINN approach. Figure \ref{FigN1} presents a schematic diagram of this inversion-based framework.
We have employed deep feed-forward neural network architectures throughout this work, using hyperbolic tangent activation functions and no additional regularization. 

The proposed inversion-based strategy facilitates a more comprehensive and systematic exploration of the solution landscape by shifting the focus from predicting solutions for fixed parameters to inferring parameters from fixed solutions. One of the key advantages of this approach is its ability to uncover rich dynamical phenomena, such as multi-stability and bifurcations. Often elusive in traditional forward simulations, these features can be more readily detected when scanning the system's behaviour across a continuum of potential equilibrium points $u_{*}$, rather than across a fixed set of parameter values. Moreover, this method allows us to identify critical thresholds for catastrophic shifts (CTCS) in the parameter $\lambda$. Specifically, we focus on detecting local extrema-minima and maxima of $\lambda$ that correspond to equilibrium points within the interior of the considered $u_{*}$-range. Thresholds can be identified by solving the equation \(\frac{d\lambda}{du_{*}} = 0\). We can determine these thresholds because during network training, we use a deep neural network (DNN) and leverage the automatic differentiation capabilities provided by machine learning libraries, such as PyTorch in Python. This enables the detection of local extremum values of $\lambda$, which mark bifurcation points indicating critical transitions in the system’s dynamics and offer valuable insights into its stability structure. We demonstrate the effectiveness of this approach through a series of illustrative examples, highlighting how the DNN-based inversion technique captures complex behaviours and delineates regions of parameter space associated with qualitative shifts in system dynamics.

\subsubsection*{Ecological examples}

A single-equation model in ecology is a mathematical framework for describing the population dynamics of a single species in isolation from others. These equations capture how a population grows or declines under different factors such as birth and death rates, resource availability, and environmental constraints. One of the simplest and most well-known examples is the logistic growth model \cite{fisher1937, kolmogoroff1988}, which incorporates carrying capacity, the maximum population size an environment can sustain. However, this classical model does not exhibit critical transitions or sudden population-level shifts. To capture such phenomena, a minimal model that reveals folded behaviour and potential regime shifts in population dynamics can be described as \cite{scheffer2001}:
\begin{equation}{\label{SSME}}
    \frac{du}{dt} = \alpha - \beta u + rg(u),
\end{equation}
with a non-negative initial condition $u(0) = u_{0}$. In this setting, the parameter $\alpha$ represents an environmental influence that promotes the accumulation of $u$, while $\beta$ defines the rate at which $u$ diminishes within the system. The parameter $r$ regulates the recovery of $u$, governed by a function $g$ that depends on $u$ itself. For instance, consider a function $g(u)$ that rises sharply near a threshold value $h$, such as the Hill function: $$g(u) = \frac{u^{p}}{u^{p}+h^{p}},$$ where the exponent $p$ controls how steeply the function transitions around the threshold $h$.

In ecological examples, if $u$ denotes the extent of barren soil, $\alpha$ can be considered vegetation loss, $\beta$ as the recolonization of degraded land by plants, and $r$ as erosion caused by wind and surface runoff. In aquatic ecosystems such as lakes, $u$ may represent the concentration of nutrients bound in phytoplankton, contributing to turbidity. In this context, $\alpha$ corresponds to nutrient loading from external sources, $\beta$ to the nutrient removal rate, and $r$ to the internal recycling of nutrients within the system.

\begin{figure*}
\centering
\includegraphics[width=0.9\textwidth]{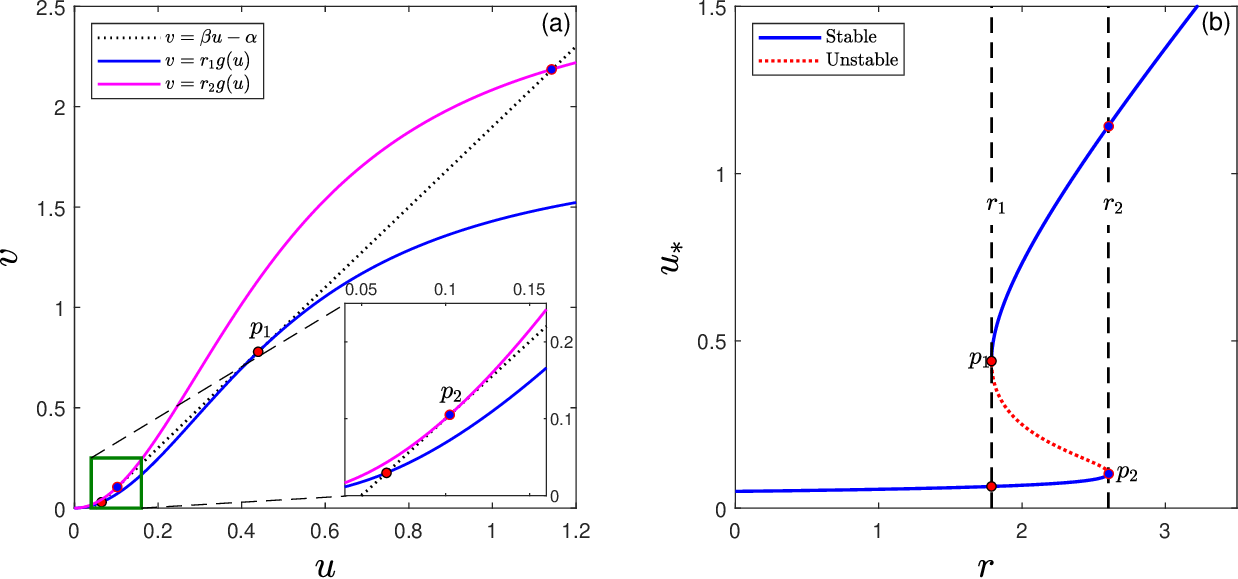}
\caption{(Color online) (a) Plots of the functions (\ref{F1}) for two different values of $r$: $r_{1} = 1.7869$ and $r_{2} = 2.6049$. (b) Bifurcation diagram of the system (\ref{SSME}) with respect to the parameter $r$. }\label{FigA1}
\end{figure*}

The model under consideration (\ref{SSME}) has a single equilibrium point at $u=\alpha/\beta$ when $r = 0$. In addition, to determine the equilibrium points of the system (\ref{SSME}), we set the right-hand side of the differential equation to zero, yielding the algebraic condition
$$\alpha - \beta u + rg(u) = 0\Rightarrow \beta u - \alpha = rg(u).$$
This equation characterizes the equilibrium states as the solutions to a balance between the linear function $\beta u - \alpha$ and the nonlinear function $rg(u)$. Therefore, the equilibrium points correspond to the intersection points, located in the first quadrant, of the following two curves:
\begin{equation}{\label{F1}}
    v = \beta u-\alpha~~~\mbox{and}~~v = rg(u).
\end{equation}
This shows that the system (\ref{SSME}) may exhibit multiple equilibrium points if the condition $rg'(u)>\beta$ is satisfied. Figure \ref{FigA1}(a) illustrates these curves for the parameter values $\alpha = 0.1$, $\beta = 2$, $h = 0.5$, $p=2$, and two different values of $r$ ($r_{1} = 1.7869$ and $r_{2} = 2.6049$).
As shown in the figure, increasing the value of $r$ results in an upward shift of the nonlinear curve $v = rg(u)$. Within the interval $r_{1} < r < r_{2}$, the two curves intersect at three distinct points in the first quadrant. This indicates that the system (\ref{SSME}) admits three equilibrium points in this range, suggesting the possibility of bistability or more complex dynamical behaviour. In contrast, when $r$ lies outside this interval, the curves intersect at exactly one point, leading to a unique equilibrium. Specifically, for $0 < r < r_{1}$, the point $p_{1}$ is absent, and for $r > r_{2}$, the point $p_2$ disappears. These transitions in the number of equilibria as $r$ varies point to the occurrence of saddle-node bifurcations at $r = r_{1}$ and $r = r_{2}$, where pairs of equilibrium points are either created or annihilated. Therefore, the qualitative behaviour of the system depends sensitively on the parameter $r$, which modulates the interaction strength in the nonlinear term $rg(u)$.

Next, we compare the equilibrium points of the model (\ref{SSME}) obtained using the traditional numerical approach and the DNN-based inversion technique. Figure \ref{FigA1}(a) presents the bifurcation diagram generated by the traditional method, where we compute all possible equilibrium points of the system for each fixed value of the parameter $r$ over the range $[0, 3.5]$. It shows that two saddle-node bifurcations occur with increasing the value of $r$ where two equilibria (one saddle and the other stable) appear at $r=r_{1}$ and two equilibrium points disappear at $r=r_{2}$. 

\begin{figure*}
\centering
\includegraphics[width=0.9\textwidth]{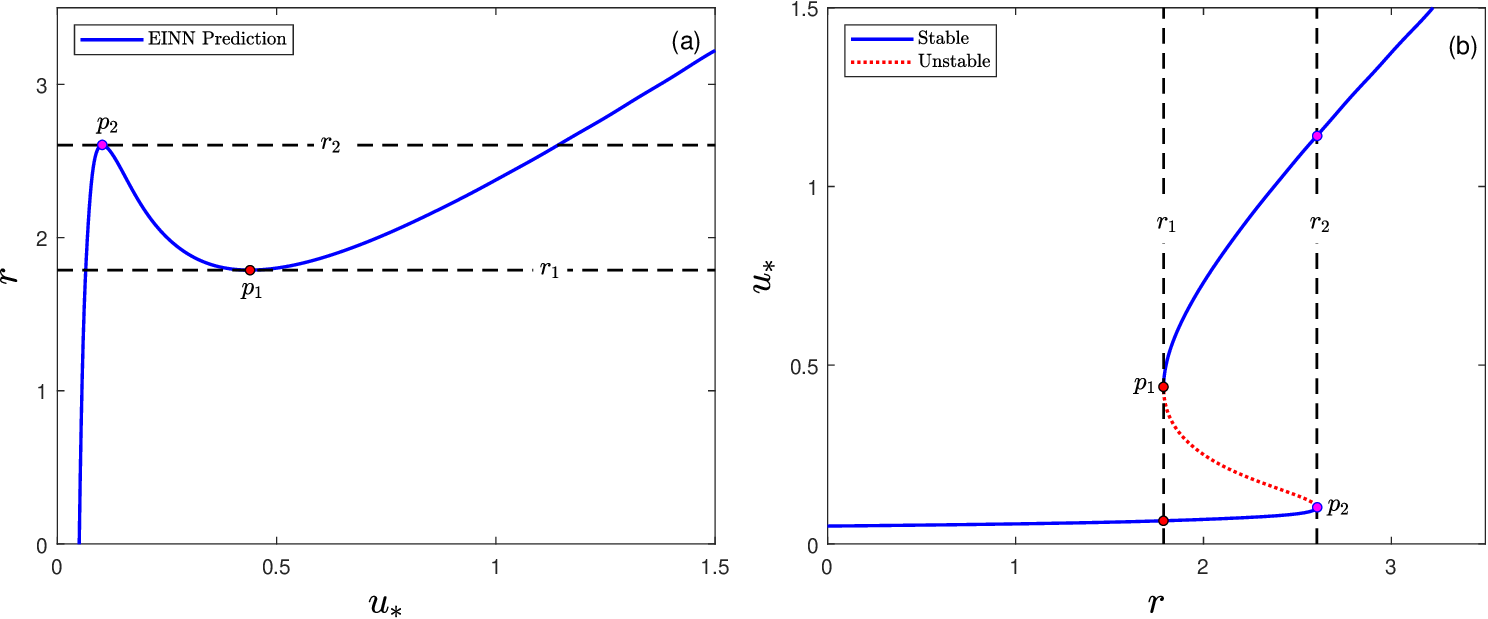}
\caption{ (Color online) Prediction of potential catastrophic shifts in the system (\ref{SSME}) using the EINNs approach. (a) Bifurcation diagram as predicted by EINNs. (b) Reoriented version of (a) with axes interchanged, overlaid with the linear stability behaviour of the equilibrium points to highlight stability transitions.} 
\label{FigB1}
\end{figure*}

In contrast, we apply the DNN-based EINNs approach, and we specify an input range for the equilibrium points, namely $[0, 1.5]$, and employ a DNN to estimate the corresponding values of $r$. The DNN architecture consists of four hidden layers, each containing ten nodes, and is trained using the Adam optimizer with a learning rate of $10^{-3}$. We train the network by minimizing the mean squared residual error as defined in equation (\ref{SEOPTM}) and stop the training when the MSE falls below $2\times 10^{-10}$. To ensure physical feasibility, we discard any negative predicted values of $r$. Figure \ref{FigB1}(a) shows the approximation of the values of $r$ obtained by the EINN method.
We interchange the axes and get the bifurcation diagram with respect to $r$ [see Fig. \ref{FigB1}(b)]. The resulting bifurcation diagram from the DNN-based approach shows good agreement with that obtained using the traditional method [see Figs. \ref{FigA1}(b) and \ref{FigB1}(b)], validating the effectiveness of the inversion technique.

Moreover, the EINNs approach enables the identification of local minima and maxima in the predicted values of $r$, and they occur at $\frac{dr}{du_{*}} = 0$, which correspond to potential critical thresholds for bifurcation. In this example, the EINNs method predicts that such local minima and maxima occur at $p_{1} = 0.4395$ and $p_{2} = 0.1025$, and their values are $r_{1} = 1.7869$ and $r_{2} = 2.6049$ [see Fig. \ref{FigB1}(a)]. These predictions align well with the expected transition points in the system through the traditional approach, demonstrating the capability of the DNN method not only to approximate equilibrium behaviour accurately but also to capture critical dynamical features of the model. Furthermore, this approach can identify a larger number of equilibrium points, if they exist, for certain parameter values, and can also predict thresholds for such critical transitions. This highlights the potential of deep learning-based inversion as a powerful tool for analyzing nonlinear dynamical systems.

Next, we study another example in ecological systems that captures the essential features of ecosystems prone to regime shifts. The model is grounded in deterministic nonlinear differential equations, representing the temporal evolution of key ecological variables such as biomass, nutrient levels, or species densities. The nonlinearities in the model- often arising from feedback mechanisms, saturating responses, or competitive interactions- give rise to bifurcation structures that can support multiple equilibrium points. We consider the non-dimensionalized model as \cite{may1977}: 
\begin{equation}{\label{SSME1}}
    \frac{du}{dt} = u(1-u) - \frac{\beta u^{2}}{u^{2}+\alpha^{2}},
\end{equation}
with a non-negative initial condition $u(0) = u_{0}$. Our goal is to identify the conditions under which these equilibria coexist and to characterize the thresholds at which the system transitions between them. We examine how parameter changes can lead to saddle-node bifurcations. Through the EINNs approach, we map out the phase space of the system and delineate regions of mono and multistability. This framework provides insight into the critical breakpoints that govern ecosystem transitions and highlights the potential for early warning indicators based on the geometry of the system’s attractors. 

The equilibrium points of the system (\ref{SSME1}) satisfy the nonlinear equation 
\begin{equation*}
    u(1-u) - \frac{\beta u^{2}}{u^{2}+\alpha^{2}} = 0 \Rightarrow u(1-u) = \frac{\beta u^{2}}{u^{2}+\alpha^{2}}.
\end{equation*}
This implies that the equilibria are the intersection points of the two curves
\begin{equation}{\label{NE2}}
    v= u(1-u)~~~\mbox{and}~~v =\beta q(u),
\end{equation}
where $q(u) = u^{2}/(u^{2}+\alpha^{2})$. In line with the approach of May \cite{may1977}, we fix the parameter value at $\alpha = 0.1$ to explore the system's dynamical behaviour. We then treat $\beta$ as the bifurcation parameter, allowing us to investigate how variations in $\beta$ influence the stability and qualitative structure of equilibria. This setup enables the identification of critical thresholds and potential regime shifts in the system, which are essential for understanding the onset of complex dynamical phenomena such as bistability or critical transitions. We employ the EINNs approach to identify potential critical transitions in the system.
\begin{figure*}[ht!]%
\centering
\includegraphics[width=0.9\textwidth]{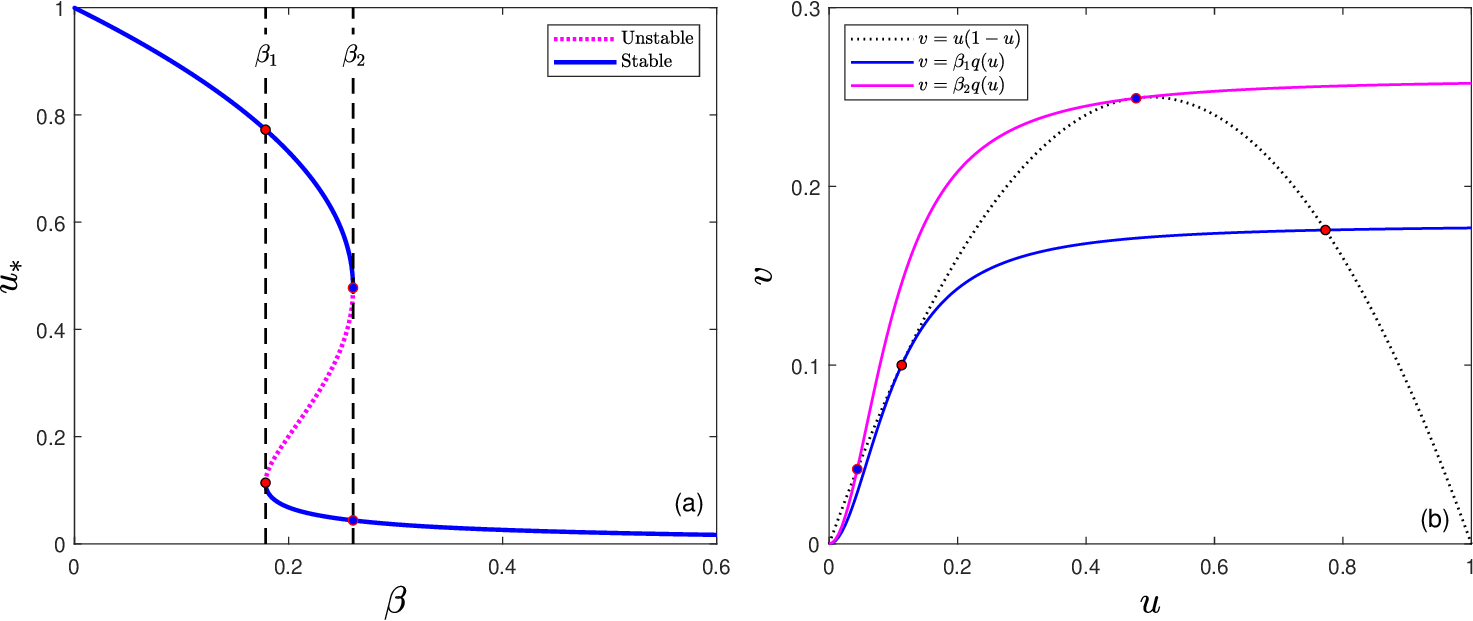}
\caption{ (Color online) (a) Bifurcation diagram and critical transition thresholds ($\beta_{1}$ and $\beta_{2}$) of system (\ref{SSME1}) derived using the EINNs method, and (b) plots of the curves from equation (\ref{NE2}) corresponding to two selected critical threshold values.} 
\label{FigD1}
\end{figure*}
Figure \ref{FigD1}(a) illustrates the bifurcation diagram of the system's attractors, with $\beta_{1}$ and $\beta_{2}$ marking the critical thresholds where qualitative changes in system dynamics occur, as determined by the EINNs method. To validate these thresholds, we examine the system's behavior by plotting the curves defined in equation (\ref{NE2}) for these two critical values of $\beta$ in Fig. \ref{FigD1}(b). The resulting plots confirm the predictions: the system exhibits the appearance and disappearance of equilibria at the critical values of $\beta$, and possesses three equilibria when $\beta$ lies between $\beta_{1}$ and $\beta_{2}$. In contrast, only a single equilibrium exists for the system when the parameter value of $\beta$ is outside the interval $[\beta_{1}, \beta_{2}]$. We further apply linear stability analysis to the equilibria across the entire range $\beta \in [0, 0.6]$, and the results are indicated in Fig. \ref{FigD1}(a). This analysis reveals that the system can undergo critical shifts as the parameter $\beta$ crosses the thresholds on either side of the interval $[\beta_{1}, \beta_{2}]$. This finding supports the robustness of the EINNs approach in accurately capturing the system’s bifurcation structure and critical transition dynamics.

\subsection{Models based on two coupled equations}

Mathematical models play a pivotal role in understanding complex biological systems by capturing essential interactions among components of interest. Among the diverse modelling approaches, two-equation models stand out as minimal yet insightful representations of biological processes involving the interaction between two key variables \cite{mumby2007}. These models are often formulated as systems of ODEs or partial differential equations (PDEs), depending on whether spatial effects are considered. In ecology, the classical Lotka-Volterra equations are a canonical example of a two-equation model describing the interaction between a predator and its prey \cite{lotka1925, volterra1926}. In theoretical neuroscience, the FitzHugh-Nagumo model simplifies the Hodgkin-Huxley model to describe neuronal excitability \cite{fitzhugh1961}. Two-equation models are common in compartmental epidemiology, such as the SIS model for infectious diseases, where recovered individuals become susceptible again \cite{hethcote2000}. In this work, we consider a general two-equation ODE model that takes the form:
\begin{equation}{\label{SOE2V}}
    \begin{aligned}
    \frac{du}{dt} &= f(u,v;\lambda),\\
    \frac{dv}{dt} &= g(u,v;\lambda),
\end{aligned}
\end{equation}
with the initial conditions $u(0) = u_{0}$ and $v(0) = v_{0}$, and the system is parametrized by $\lambda$. 

The critical transitions in the system (\ref{SOE2V}) refer to abrupt qualitative changes in the system's behaviour as the parameter $\lambda$ is varied, and they often correspond to bifurcations - points where the number of equilibrium states changes. As $\lambda$ crosses a critical value, the system may shift from one stable state to another. These critical transitions can emerge through various bifurcation scenarios, such as saddle-node bifurcation, which leads to the sudden disappearance of a stable fixed point, leaving the system to evolve toward a different attractor, often far from the original state. In the context of this system, such transitions manifest as sudden jumps in the values of $(u(t),v(t))$ after slow, continuous changes in $\lambda$. We find these critical shifts that occur through the generation or the disappearance of a stable equilibrium point in a system by identifying all equilibrium points across a range of $\lambda$ values. 

The traditional approach for analyzing equilibrium points involves solving the nonlinear equations $f(u,v;\lambda)=0$ and $g(u,v;\lambda)=0$ for each fixed value of the parameter $\lambda$, and then plotting all such solutions to visualize how the number and nature of equilibria evolve with respect to $\lambda$. This method effectively captures bifurcations by tracking how equilibria appear, disappear, or change stability as $\lambda$ varies. However, in this work, we adopt the EINNs approach to identify equilibrium points of the system \eqref{SOE2V} across a range of $\lambda$ values. Rather than solving for equilibria pointwise in $\lambda$, our EINNs method seeks values of $\lambda$ and $v$ that, together with a prescribed $u_{*}$, satisfy the equilibrium conditions simultaneously. Specifically, we consider a given set of candidate values $\{u_{*}^j\}_{j=1}^N$ and aim to identify corresponding values $(v_{*}^j, \lambda^j)$ such that the conditions
$$f(u_{*}^j, v_{*}^j; \lambda^j) = 0 \quad \mbox{and}~~ g(u_{*}^j, v_{*}^j; \lambda^j) = 0$$
are satisfied for each $j = 1, 2, \ldots, N$. Alternatively, one may choose a set of candidate values $\{v_{*}^j\}_{j=1}^N$ and aim to identify corresponding values $(u_{*}^j, \lambda^j)$ such that the above conditions hold. In both cases, the resulting bifurcation diagram will be the same. 

In the EINNs approach, we train a DNN to minimize the residuals of the governing equations over the sample points, effectively learning the structure of equilibrium solutions as a function of the chosen $u_{*}$ values. The objective function for training is the mean squared residual error, defined as:
\begin{equation}\label{SEOPTM1}
\text{MSE} = \frac{1}{N} \sum_{j=1}^{N} \min_{v} \left\{ |f(u_{*}^{j}, v; \lambda^{j})|^2 + |g(u_{*}^{j}, v; \lambda^{j})|^2 \right\}.
\end{equation}
This error measures the extent to which the neural network-generated $\lambda^{j}$ values for each fixed $u_{*}^j$ satisfy the equilibrium conditions. In addition, the values of $v_{*}^{j}$ can be identified by searching for the arguments of the minima in (\ref{SEOPTM1}) for each $j$ ($j=1,2,\dots, N$). 

\subsubsection*{Examples from neurodegenerative disease studies}

Alzheimer’s disease (AD) is a progressive neurodegenerative disorder characterized by cognitive decline and extensive neuronal damage, with hallmark pathological features including extracellular accumulation of amyloid-beta (A$\beta$) plaques and disrupted intracellular calcium (Ca$^{2+}$) homeostasis. Emerging experimental evidence suggests that A$\beta$ and Ca$^{2+}$ are engaged in an unfavourable positive feedback loop, wherein A$\beta$ promotes excessive Ca$^{2+}$ influx, and elevated intracellular Ca$^{2+}$ further enhances A$\beta$ production and aggregation. Understanding the dynamics of this mutual reinforcement is crucial to elucidating disease progression and identifying intervention points. Deterministic mathematical models offer a powerful framework to capture and analyze these complex biochemical interactions by formulating them as systems of differential equations grounded in biophysical and biochemical principles. In this context, we consider a deterministic model to quantify the interplay between A$\beta$ and Ca$^{2+}$ concentrations over time, incorporating key processes such as A$\beta$ production, aggregation, degradation, Ca$^{2+}$ influx through membrane channels, and Ca$^{2+}$ clearance mechanisms as \cite{de2013, pal2025}:
\begin{equation}{\label{ABCI1}}
    \begin{aligned}
        \frac{du}{dt} &= a_{1}+a_{2}f_{1}(v) - k_{1}u,\\
        \frac{dv}{dt} &= \frac{1}{\epsilon}(b_{1}+b_{2}g_{1}(u) - k_{2}v),
    \end{aligned}
\end{equation}
with non-negative initial conditions. Here, $u(t)$ and $v(t)$ denote the concentrations of A$\beta$ and Ca$^{2+}$, respectively. The parameter $a_{1}$ represents the basal production rate of A$\beta$, while $k_{1}$ is its degradation rate. The term $a_{2} f_{1}(v)$, with $f_{1}(v) = v^{2}/(v^{2} + \alpha^{2})$, introduces a nonlinear positive feedback from Ca$^{2+}$ to A$\beta$ production, modelled by a Holling type-III function to capture saturation effects at high Ca$^{2+}$ concentrations. For the Ca$^{2+}$ dynamics, $b_{1}$ denotes the basal influx rate and $k_{2}$ is the natural clearance rate. The term $b_{2}g_{1}(u)$, with $g_{1}(u) = u$, models the positive regulation of Ca$^{2+}$ influx by A$\beta$. The small parameter $\epsilon$ ($0 < \epsilon\leq 1$) distinguishes the fast dynamics of Ca$^{2+}$ (seconds) from the much slower accumulation of A$\beta$ (years), enabling the study of multi-timescale interactions. This model framework sets the stage for exploring conditions under which pathological feedback may induce sustained elevation in both A$\beta$ and Ca$^{2+}$ levels, potentially offering insights into tipping points in disease progression and the design of therapeutic control strategies.

Analyzing the model’s behaviour under various parameter regimes allows one to investigate the conditions under which the system transitions from a healthy equilibrium to a pathological state characterized by sustained high levels of both A$\beta$ and Ca$^{2+}$. This study enables the exploration of feedback-driven bistability or threshold effects, which may underlie the irreversible progression of AD. Ultimately, this approach contributes to a deeper theoretical understanding of disease mechanisms and aids in the design of targeted therapeutic strategies aimed at disrupting the A$\beta$-Ca$^{2+}$ feedback loop. 

\begin{figure*}[ht!]%
\centering
\includegraphics[width=0.85\textwidth]{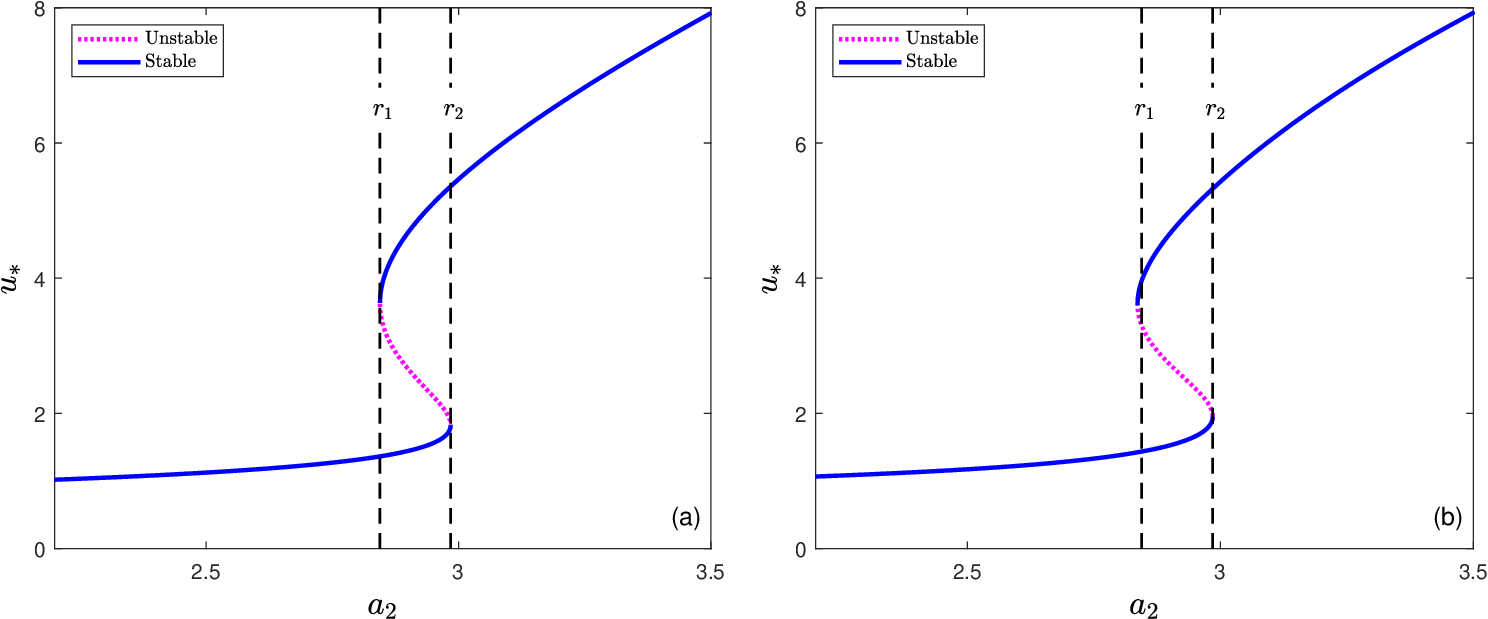}
\caption{ (Color online) Comparison of equilibrium point estimates for system (\ref{ABCI1}) obtained via (a) traditional and (b) EINNs approaches.} 
\label{FigC1}
\end{figure*}

In this example, we adopt the fixed parameter values from \cite{de2013, pal2025} as follows: $a_{1} = 0.25$, $\alpha = 1.0$, $k_{1} = 0.35$, $b_{1} = 0.11$, $b_{2} = 1.0$, and $k_{2} = 5.0$, treating $a_{2}$ as the bifurcation parameter. It is important to note that the parameter $\epsilon$ does not influence the determination of the system’s equilibrium points in equation (\ref{ABCI1}). As a result, it does not affect the location of any critical transition thresholds if they exist. Figure \ref{FigC1} presents a comparative analysis of the bifurcation diagrams generated using the traditional method and the EINNs approach.
The comparison reveals that the EINN-based method yields results in strong agreement with traditional bifurcation analysis, accurately capturing key qualitative features and critical transitions in the system’s behaviour. This consistency underscores the potential of the EINN framework as a reliable tool for exploring nonlinear dynamical systems, particularly in scenarios where analytical techniques are challenging to apply.

\subsection{Extended systems of equations}

Real-world systems-whether ecological, biological, physical, or socio-economic- are inherently complex, involving numerous interacting components that influence one another across different timescales and spatial scales. To accurately capture the essential dynamics of such systems, it is often necessary to model them using systems of multiple coupled ODEs. Each equation typically represents a distinct variable or process, such as population densities, nutrient concentrations, or behavioural responses, and their interactions are governed by nonlinear terms that reflect feedback, competition, cooperation, or adaptation. As the number of interacting variables increases, so does the capacity of the model to reproduce rich and sometimes unexpected dynamics, including oscillations, chaos, multistability, and abrupt regime shifts. Therefore, moving beyond simple two-variable models is essential for studying and predicting the behaviour of complex systems under realistic conditions.

A classic example is the Rosenzweig–MacArthur predator-prey-resource model, a three-variable system that can exhibit catastrophic collapse of predator populations due to changes in resource availability or predation rates \cite{rosenzweig1971}. In addition, models of nutrient–phytoplankton– zooplankton dynamics in aquatic ecosystems have shown sudden regime shifts between clear and turbid water states \cite{scheffer2001}. These systems are typically governed by nonlinear interactions and multiple feedback loops, where small parameter changes can lead to large qualitative changes in system behaviour. We consider these systems of equations as
\begin{equation}{\label{SOEM}}
    \begin{aligned}
        \frac{du_{1}}{dt} =& f_{1}(u_{1},\ldots, u_{n};\lambda),\\
        & \vdots \\
        \frac{du_{n}}{dt} =& f_{n}(u_{1},\ldots, u_{n};\lambda),
    \end{aligned}
\end{equation}
with appropriate initial conditions and $n\geq 1$. The parameter $\lambda$ plays a central role in shaping the system's dynamics. To explore the equilibrium behaviour of the system described by equation \eqref{SOEM}, we utilize the EINNs framework, which is designed to efficiently identify steady-state solutions of the form $(u_{1*}^{j},\ldots,u_{n*}^{j})$ by systematically varying the bifurcation parameter $\lambda$. This methodology extends our prior strategies developed for lower-dimensional systems, such as one- and two-equation models.

Our approach begins by prescribing a set of $N$ candidate values for the first component of the steady state, denoted $\{u_{1*}^{j}\}_{j=1}^{N}$. For each fixed $u_{1*}^{j}$, the goal is to determine the associated parameter value $\lambda^{j}$ and the remaining components $(u_{2*}^{j}, \ldots, u_{n*}^{j})$ such that the full system satisfies the equilibrium conditions:
$$f_{1}(u_{1*}^{j},\ldots, u_{n*}^{j};\lambda^{j}) = 0, ~\ldots, f_{n}(u_{1*}^{j},\ldots, u_{n*}^{j};\lambda^{j}) = 0,$$
for each $j=1,\ldots,N$. To solve this problem, we train a DNN within the EINNs framework to minimize the residuals of the reaction terms across all sampled points. Specifically, the network is optimized to approximate the equilibrium manifold by reducing the total deviation from the steady-state conditions at each selected value of $u_{1*}^j$. The learning objective is formalized through a mean squared error (MSE) loss function defined as:
\begin{equation}\label{MEOPTM1}
\text{MSE} = \frac{1}{N} \sum_{j=1}^{N} \min_{u_{2},\ldots,u_{n}} \left\{ \sum_{k=1}^{n} |f_{k}(u_{1*}^{j},u_{2},\ldots,u_{n}; \lambda^{j})|^2 \right\}.
\end{equation}
This formulation enables the network to approximate the set of steady states by implicitly learning the functional dependencies between the state variables and the bifurcation parameter. As a result, the EINNs method offers a scalable and flexible alternative to traditional numerical continuation techniques, particularly in higher-dimensional settings where direct computation of equilibria becomes challenging.

\section{Conclusions}{\label{SE4}}

In this work, we introduce a novel DNN framework, the EINNs approach, for solving inverse problems in dynamical systems by reversing the conventional paradigm. Instead of fixing parameters and seeking the resulting equilibrium solutions, EINNs begin with candidate equilibrium states and determine the corresponding parameters that validate them. This inversion-based strategy enables a more targeted and efficient investigation of parameter spaces, which is particularly beneficial in nonlinear systems with complex or bifurcating equilibria where direct parameter scans are computationally intensive or insufficiently informative. 

A major strength of the EINN method is its ability to approximate the inverse mapping from equilibrium states to parameter values using a DNN trained to minimize residuals of the governing equations. This data-driven approach circumvents the need for repeated numerical solvers and allows generalization across a wide range of inputs. It has the capacity to uncover diverse solutions that may be overlooked by conventional methods, especially in systems exhibiting multistability or abrupt regime shifts. By directly focusing on candidate equilibria and learning the associated parameter configurations, EINNs can detect critical thresholds, such as bifurcation or tipping points, with enhanced precision. Furthermore, the EINNs approach is especially powerful in high-dimensional systems and those governed by implicit or analytically intractable relationships. 

Despite its advantages, the EINN approach comes with certain limitations. Its performance is sensitive to the sampling strategy for candidate equilibria, such as when the range of the given candidate equilibria is too wide. Additionally, the critical threshold cannot be detected if regime shifts occur outside the training data. Moreover, neural networks require careful design, training, and validation to prevent issues such as overfitting, vanishing gradients, or convergence to local minima. 

The versatility of EINNs opens doors to several impactful application areas. In ecology, they can be used to infer environmental conditions leading to population equilibria or tipping points in ecosystems. In systems biology, EINNs may help identify biochemical reaction rates that support steady-state concentrations in gene regulatory or metabolic networks. In physics and engineering, they can assist in inverse design problems, such as determining material or boundary parameters that yield specific equilibrium fields. Moreover, in neuroscience, EINNs can offer insights into synaptic or neuronal parameters underlying stable activity patterns associated with healthy or pathological brain states. These diverse applications demonstrate the potential of EINNs as a general-purpose tool for equilibrium inference in complex systems.

\section*{Acknowledgements}
The authors thank the NSERC and the CRC Program for their support. RM also acknowledges the support of the BERC 2022-2025 program and the Spanish Ministry of Science, Innovation and Universities through the Agencia Estatal de Investigacion (AEI) BCAM Severo Ochoa excellence accreditation SEV-2017-0718. This research was partly enabled by support provided by SHARCNET and the Digital Research Alliance of Canada.

\bibliography{References}

\end{document}